\documentclass[conference]{IEEEtran}
\usepackage{ifpdf}

\usepackage{cite}

\ifCLASSINFOpdf
  \usepackage[pdftex]{graphicx}
  \usepackage{caption}
\else
\fi
\begin{document}
%
\title{Reinforcement Learning Applied to Single Neuron}

\author{\IEEEauthorblockN{Zhipeng Wang}
\IEEEauthorblockA{Department of Chemistry\\
Rice University\\
Houston, Texas 77005\\
Email: zw12@rice.edu}
\and
\IEEEauthorblockN{Mingbo Cai}
\IEEEauthorblockA{Department of Neurosciences\\ Baylor College of Medicine \\
Houston, Texas, 77005\\
Email: mingbo.cai@gmail.com}}


%


\maketitle

\begin{abstract}
This paper extends the reinforcement learning ideas into the multi-agents system, which is far more complicated than the previously studied single-agent system. We studied two different multi-agents systems. One is the fully-connected neural network consists of multiple single neurons. Another one is the simplified mechanical arm system which is controlled by multiple neurons. We suppose that each neuron is like an agent and it can do Gibbs sampling of the posterior probability of stimulus features. The policy is optimized in a way that the cumulative global rewards are maximized. The algorithm for the second system is based on the same idea but we incorporate the physics model into the constraints. The simulation results show that for the first system our algorithm converges well. For the second system it does not converge well in a reasonable simulation time length. In summary, we took the initial endeavor to study the reinforcement learning for multi-agents system. The computational complexity is always an issue and significant amount of works have to be done in order to better understand the problem. 
\end{abstract}


%
\IEEEpeerreviewmaketitle

\section{Introduction}
Reinforcement Learning (RL) is a sub-field of machine learning, by which the agent interacts with the environment to dynamically maximize its long-term expected rewards (1, 2).  It has wide applications in a variety of emerging fields such as robotics, artificial intelligence, brain and cognitive sciences and neurosciences. In reality, Animals make decisions by the neural system. The neural system consists of single neurons. Each of them has some available actions, the collective action of the entire neural network determines the action of the system (animals, robots etc.) and by which the system makes decisions. The task for reinforcement learning algorithm is trying to optimize the policy under which the system takes actions to maximize its cumulative long-term rewards. 
There are a lot of exciting research works already done regarding reinforcement learning algorithms, from both the theoretical algorithmic design/analysis and practical applications of RL (1-6). However, most of the previous research works are focused on single agent policy optimization, in which there is only one agent interacting with the environment (4-6); in real world, the system makes decisions through the policy optimization of the multi-agents network. For example, animals make decisions by the neural network, which is the fully-connected neural system consists of single neurons. Each single neuron is like an agent, they are correlated and interacting with each other through the network. The system is making decisions by optimizing the cumulative global rewards of the entire networks. The tricky part of this kind of problems is that we don’t really know how the different agents are correlated, and how to define a global reward function for the entire network in order to do the policy iteration. Also as the network becomes bigger, there are large numbers of single neurons connected via different correlation functions, if we don’t choose the appropriate models and parameters, the global policy iteration might not be necessarily converge. There is also a concern about the computational complexity, as the global reward calculation is not as easy as that in single-agent problems. 
In this paper, we want to take the endeavor to study the multi-agents reinforcement learning problem. There are two contributions that we made in this paper. The first contribution is that we developed an effective learning algorithm to train the simple neuron network to reach the desired goal; the algorithm is based on the idea of reinforcement learning. We suppose that each neuron plays as a single agent, which can do Gibbs sampling of posterior probability of stimulus features. The computer simulation shows that the algorithm converges well for the system we tested. The second contribution of the paper is that we extended the basic ideas of multi-agents learning algorithm into a more practical application in robotics, which is about positioning the mechanical arm. The basic idea is that when the robot sees an object, it would use its arm to reach that object. In order to train the robot to reach the right position, we need to apply the multi-agent RL algorithm to maximize the cumulative rewards of the mechanical arm actions. In our model, we defined a multi-agents system and physics model which determine the motion of the mechanical arm. Although the algorithm does not converge well, it provides a new way to think about the motion-planning problem in robotics, in which the RL techniques could have huge potential applications. 
The paper is organized as follow. Section 2 is about the detailed algorithms and methodologies that we developed. Section 3 shows the simulation results, and Section 4 is the conclusion and future works. The paper ends with acknowledgement and references. 


 

\section{Algorithms}
The merit of the algorithm resides in applying reinforcement learning to each individual neuron in a network. Specifically, our idea is inspired by the gradient ascent approach for reinforcement learning. 
A neuron is considered as an agent. It receives inputs from other neurons and external inputs. Its output is action potentials. For simplicity, we model each neuron as single compartment which has a membrane potential. Spike count is generated by a Poisson process, of which the firing rate is a sigmoid function of the membrane potential. The neurons have leaky current to regress their membrane potentials toward resting potential. A spike of a neuron generates either an increase or decrease of membrane potential in the neuron receiving input from it. The size of change depends on the synaptic weights between them. \\

The combination of a neuron’s synaptic weights reflects its policy to interact with the environment. The only way that a neural network can improve its performance is by adjusting the synaptic weight of individual neurons. The improvement of synaptic weights is achieved by two processes: sampling weight distribution and updating weights by gradient descent. Instead of assuming the synaptic weights are fixed for a neuron, we model the weights as Gaussian process with auto-correlation. In other words, they fluctuate around a center value. This not only reflects the stochastic property of biological synapses, but also achieves sampling of a distribution of synaptic weights. This also corresponds to a stochastic policy for an agent. \\

In one epoch, a neuron’s synaptic weights sample through Gaussian processes:\\
\begin{equation}
w_t = \alpha w_{t-1} + (1- \alpha) (w_0 + \epsilon)
\end{equation}
Where $0< \alpha <1$, $w_0$ is the expected weight for a specific synaptic connection. \\
The global reward is defined by some metrics of the firing activity of the whole or a part of the network. The reward for a neuron is the combination of this global reward and some metric of its only activity. For example, firing too many action potential consumes a lot of energy. Therefore one reasonable reward related to the neuron itself can be -1 for each of its spike. \\

After the reward time course is available for a neuron on an epoch, the estimated value associated with each sampled synaptic weight during the epoch is defined by the discounted reward after that time point: \\
\begin{equation}
V(w_{t_i}) = \Sigma_{t=t_i}^{T} \gamma ^{t-t_i} v_t
\end{equation}
Where $v_t$ is the reward at time t. \\
A good synaptic weight should generate larger long term reward. Therefore, at the next epoch, the mean of the weight $w_0$ should move towards a better weight:\\
\begin{equation}
w_0 \leftarrow w_0 + \beta \frac{\Sigma_{t=0}^{T} v(w_t)w_t}{\Sigma_{t=0}^{T} v(w_t)}
\end{equation}
The right hand side of Eq. 3 is in fact an estimation of the partial derivative of discounted reward over the weight. If a weight has reached a local maximum, the right side part approaches 0 asymptotically.\\

Different sampling and updating scheme can be adopted. We can allow only one or a few neuron to sample and update in an epoch or all the neurons to sample and update simultaneously. The former may be slow in convergence, whereas the latter one may be vulnerable to more random walk of $w_0$ due to misattribution of reward to the wrong neurons.
\section{Simulation Results}
We first simulate a simple network of 5 neurons. All neurons are initially interconnected with random weights. Neuron 1 receives external current injection, such that its firing rate largely follows the magnitude of injected current. The time course of injected current is a sinusoidal signal with random frequency and initial phase in each epoch, such that frequency across epochs follows 1/f distribution. The global reward is defined as the dot product of spike counts between neuron 1 and neuron 2, normalized by the average of their spike counts. This reward definition achieves coincidence detection. \\

In ideal case, in order to maximize global reward, neuron 2 should have strong input weight from neuron 1 to follow its activity. Neuron 1 may also have strong input weight from neuron 2, depending on whether the external input to neuron 1 is too strong to shunt the impact of other neurons. Other neurons should decrease their input weights to reduce average response, because any single spike counts as a punishment. \\
\begin{figure}[hH]
\centering
\includegraphics[width=2.5in]{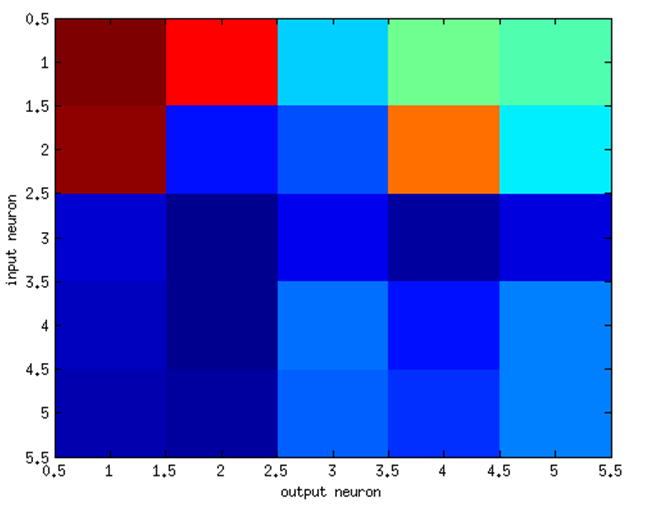}
\caption{\small Final weight of one simulation of 40000 epochs}
\end{figure}

As can be seen from Figure 1, neuron 2 indeed conserves strong input weight from neuron 1. Other neurons basically shut down input from other neurons in order to reduce their own activity. Neuron 1 is perhaps too strongly driven by external activity, such that its input weight has small influence on its activity. Therefore neuron 1 did not reduce its input from other neurons significantly. \\
\begin{figure}[hH]
\centering
\includegraphics[width=3in]{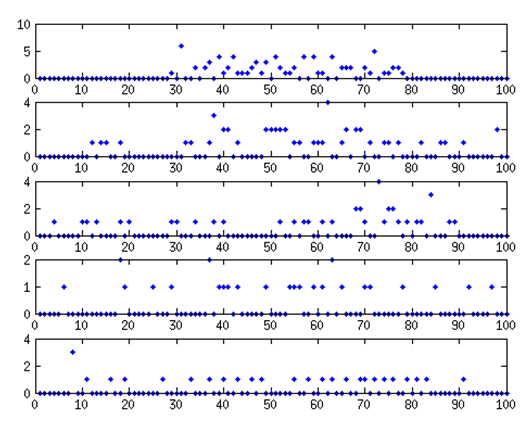}
\caption{\small Example spike count of neurons after convergence. Neuron 2 has good correlation with neuron 1}
\end{figure}

We further attempted to apply the same framework to a more complex case: motor control.\\

If two neurons in a network directly control two muscles connected to an arm, a few external neurons encode the intended angular location of the arm with Gaussian tuning curve to angles, and a few more neurons encode for observed angular location of the arm with a potential delay, can the network learn to make the arm follow the intended angle? \\

We used 20 neurons to encode intended angle, and 20 neurons to encode observed angle. 50 neurons are in the network, each of which receives inputs from all these neurons and each other. The global reward is defined as the correlation between the pattern of spike count of neurons encoding intended angle and neurons encoding observed angles. 
However, we could not get significant improvement. \\

We hypothesize the failure may be because the network is relatively large compared to the previous example. Therefore, it may be harder to climb the hill on a high dimensional profile of reward functions.  \\

A future improvement may be dividing up the reward function, and allowing sub-groups of neurons to optimize for each part of the reward function. This way, fewer neurons compete together to improve each reward function. \\

\begin{figure}[hH]
\centering
\includegraphics[width=3in]{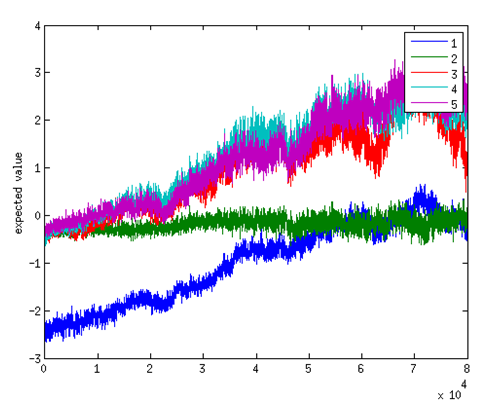}
\caption{\small The expected reward actually keeps increasing for most of the neurons}
\end{figure}

\section{Conclusion}
Reinforcement learning by gradient ascent can be applied to single neuron to allow a recurrently connected network achieve simple task. Our algorithm does not require any prior assumption of the network topology. Starting with all-connected neural network, our algorithm could be applied in a favorable way in terms of training simple networks. However, for more complicated networks the convergence time for our algorithm is unknown. Rigorous mathematical proof is needed in order to find a better reward function for more complicated neural network training.


\section*{Acknowledgment}
The authors would like to thank Dr. Richard Baraniuk for his guidance and suggestions to this work. We also would like to thank Department of Electrical and Computer Engineering at Rice University for accomodating the reinforcement learning seminar.



%

\end{document}